%% file: main.tex
\documentclass[runningheads]{llncs}

\usepackage{eccv}

\usepackage{eccvabbrv}

\usepackage{graphicx}
\usepackage{booktabs}
\usepackage{multirow}
\usepackage[accsupp]{axessibility}  

\usepackage{hyperref}

\usepackage{orcidlink}

\newcommand\blfootnote[1]{%
  \begingroup
  \renewcommand\thefootnote{}\footnote{#1}%
  \addtocounter{footnote}{-1}%
  \endgroup
}

\begin{document}

\title{Explicitly Guided Information Interaction Network for Cross-modal Point Cloud Completion} 

\titlerunning{EGIInet}

\author{Hang Xu\inst{1*} \and
Chen Long\inst{1*}\and
Wenxiao Zhang\inst{2\text{†}} \and
Yuan Liu\inst{3} \and
Zhen Cao\inst{1} \and
Zhen Dong\inst{1} \and
Bisheng Yang\inst{1}}

\authorrunning{H.~Xu and C.~Long et al.}

\institute{LIESMARS, Wuhan University \\
\email{\{190107xh, chenlong107, zhen.cao, dongzhenwhu,bshyang\}@whu.edu.cn} \and
University of Science and Technology of China \\ \email{wenxxiao.zhang@gmail.com}
\and The University of Hong Kong \\ 
\email{yuanly@connect.hku.hk}
}

\maketitle

\input{Sec/abstract}
\input{Sec/intro}
\input{Sec/rw}
\input{Sec/method}
\input{Sec/experiment}
\input{Sec/conclusion}
\section*{Acknowledgements}
This work was supported by the National Key Research and Development Program of China under Grant 2022YFB3904102.
\bibliographystyle{splncs04}
\bibliography{main}

\end{document}

%% file: Sec/abstract.tex
\begin{abstract}
\blfootnote{$^{*}$ Equal contribution \quad $^{\text{†}}$ Corresponding author}In this paper, we explore a novel framework, EGIInet (Explicitly Guided Information Interaction Network), a model for View-guided Point cloud Completion (ViPC) task, which aims to restore a complete point cloud from a partial one with a single view image. In comparison with previous methods that relied on the global semantics of input images, EGIInet efficiently combines the information from two modalities by leveraging the geometric nature of the completion task. Specifically, we propose an explicitly guided information interaction strategy supported by modal alignment for point cloud completion. First, in contrast to previous methods which simply use 2D and 3D backbones to encode features respectively, we unified the encoding process to promote modal alignment. Second, we propose a novel explicitly guided information interaction strategy that could help the network identify critical information within images, thus achieving better guidance for completion. Extensive experiments demonstrate the effectiveness of our framework, and we achieved a new state-of-the-art (\textbf{+16\% CD} over XMFnet) in benchmark datasets despite using fewer parameters than the previous methods. The pre-trained model and code and are available at \url{https://github.com/WHU-USI3DV/EGIInet}.

  \keywords{point cloud completion \and cross-modality \and multi-modal fusion}
\end{abstract}

%% file: Sec/intro.tex
\section{Introduction}
\label{sec:intro}
The extensive application scenarios and significant research value of 3D Computer Vision have garnered increasing attention. Point clouds \cite{9127813}, serving as a discrete representation of stereoscopic space, play a crucial role in various areas such as 3D reconstruction \cite{ma2018review}, scene understanding \cite{hou20193d,gu2019survey}, and autonomous driving \cite{li2020deep,cui2021deep}. However, due to inherent constraints imposed by scanning sensors, reflections and occlusions, the raw point clouds obtained from 3D scanners are often sparse, noisy, and \textbf{occluded} \cite{fei2022comprehensive,pc2pu,sparsedc}. Hence, it is necessary to conduct point cloud completion on this raw data before applying it to downstream tasks like point cloud segmentation \cite{nguyen20133d,xie2020linking} and reconstruction \cite{berger2014state,xu2021toward} and so on. To achieve this, point cloud completion emerges as a cost-effective and desirable way to restore the complete shape of the underlying surface.

Traditional point cloud completion methods \cite{dai2017shape,Huang_2020_CVPR,Tchapmi_2019_CVPR,lyu2021conditional,Xiang_2021_ICCV,zhou2022seedformer,zhu2023svdformer,xie2020grnet,zhang2020detail,Wang_2020_CVPR,yuan2018pcn,yu2021pointr,ying2023adaptive,sarmad2019rl,Wen_2021_CVPR,cui2023p2c,wang2020cascaded,9525242,wen2022pmp,xiang2021snowflakenet,Tang_2022_CVPR,li2023proxyformer,Chen_2023_CVPR,yu2023adapointr} aim to restore the complete shape from given incomplete point clouds. However, due to the inherent sparsity and unstructured nature of point clouds, learning the mapping from incomplete shapes to complete shapes solely based on point cloud data is extraordinarily challenging. As a more pragmatic option, \cite{zhang2021view} introduced the task of View-Guided Point Cloud Completion, wherein a partial point cloud is supplemented with an additional single view image to facilitate a more coherent completion.

\begin{figure}
    \centering
    \includegraphics[width=0.8\linewidth]{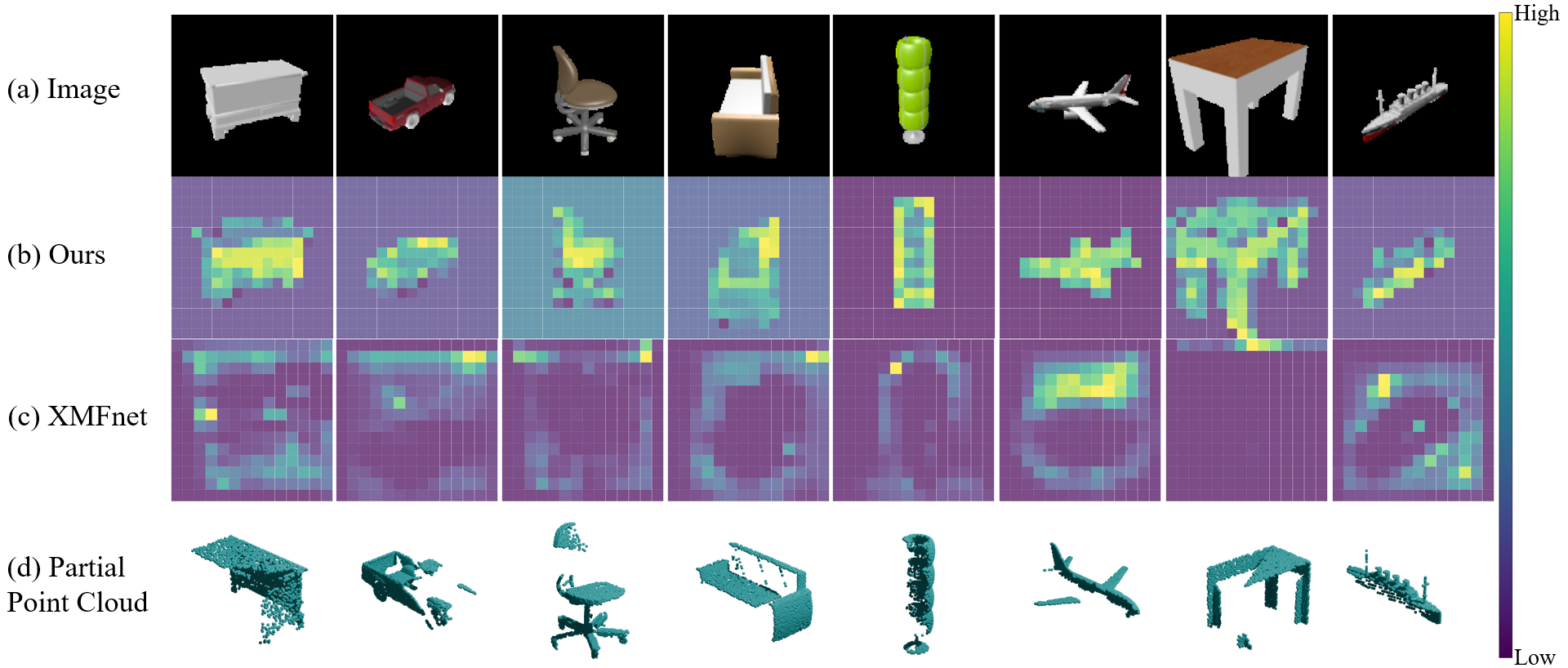}
    \caption{Cross-attention weight map projection of our method (b) and XMFnet\cite{aiello2022cross} (c). Compared with XMFnet\cite{aiello2022cross}, our method extracts clearer structural information about the images.}
    \label{fig:1}
\end{figure}

Unfortunately, while the image provides rich texture and structure information to guide the completion procedure, the inputs from different modalities also brought significant challenges for the design and training of models. To address this issue, ViPC \cite{zhang2021view} and CSDN \cite{zhu2023csdn} leverage the ideas from single-view reconstruction methods for result-level fusion with partial point clouds. However, estimating the 3D coordinates from images is an ill-posed problem \cite{han2019image}. Inspired by recent multi-modal fusion approaches, the most recent work XMFnet \cite{aiello2022cross} proposed a fusion strategy based on latent space operations that incorporate a cross-attention mechanism to conduct information fusion among multi-modal features. Nevertheless, XMFnet \cite{aiello2022cross} overlooks the inherent domain differences between inputs, and the indiscriminate stacking of cross-attention layers simply lacks explicit guidance for the process of information fusion. As shown in Fig. \ref{fig:1} (c), we visualize the attention map of image features within the cross-attention layers. It can be observed that XMFnet \cite{aiello2022cross} tends to gain the abstract global feature from the images, aiming to estimate global semantics while neglecting the inherent geometric structural characteristics of point cloud completion tasks, thus leading to sub-optimal completion outcomes.

In order to solve this problem, we rethink the fundamental nature of the View-guided point cloud completion task, and consider the most important question: How to find the critical information contained in a corresponding image and fuse it into the completion process?

To answer this question, we propose a novel completion framework named EGIInet, which identifies the critical information within images by explicitly guiding the information interaction, thus enhancing the effectiveness of single-view images in guiding the completion process. Specifically, We divide the completion process into two steps:\textbf{ Modal Alignment} and \textbf{Information Fusion}. Fig. \ref{fig:2} illustrates the whole pipeline of EGIInet.

Firstly, diverging from existing methods that use different backbone networks to extract features, we have devised a unified multi-modal feature extractor aimed at mitigating modal disparities and reducing the difficulty of subsequent information interaction.
Tokenization techniques are adopted to map data of different modalities into a unified representation, and a shared encoder structure is used to unify the learning process, thus ensuring features from different modalities are compatible in latent space. Each token feature contains the local geometry and the arrangement of the token sequence encapsulates the global structure. Through this unified encoder structure, the modal disparities among image features and point cloud features can be effectively reduced, thus promoting and simplifying the subsequent information interaction and feature fusion.


Secondly, instead of fusing the two modality features directly for completion like previous methods, we expect the network could perceive the corresponding relation between the point cloud tokens and image tokens, which could help the network figure out which image tokens are helpful for point cloud completion. To achieve this, we propose a separated information interaction process with explicit structural guidance, which is achieved by an indirect interaction network supervised by a dual-designed loss function. Through this interaction process, the structural information in the image and point cloud can be transferred to each other. Finally, we fuse these two "transferred” features with only one simple cross-attention layer for final completion. Fig. \ref{fig:1} (b) visualizes the weight map of image features within our cross-attention layer, demonstrating that our network could find the important structures for completion by performing explicit guided interaction between token features of images and point clouds, thus achieving better completion.

We conduct a comprehensive experimental evaluation of our approach on the benchmark dataset, where we achieved a \textbf{16\% improvement} over the SOTA method XMFnet \cite{aiello2022cross} in terms of the CD metric, despite utilizing fewer parameters (\textbf{9.03M} < 9.57M).


Our contribution can be summarized as follows:
\begin{itemize}
    \item We analyze the limitations of mainstream methods and propose a novel point cloud completion framework called EGIInet. It consists of a unified encoder and a novel token feature structure transfer loss to provide an explicitly guided information interaction, which could help get more reliable and better performance for the completion task.

    \item  We assess the performance between ours and other SOTA methods on some simulated and real challenging datasets. Extensive experiments show the effectiveness of our methods, our method achieves superior performance, reaching the state of the art.
\end{itemize}

%% file: Sec/rw.tex
\section{Related Work}
\label{sec:rw}
\subsection{Point cloud completion}
The pioneer work of point cloud completion is PCN \cite{yuan2018pcn}, which proposed a coarse-to-fine approach that is widely referenced in following studies \cite{liu2020morphing,Huang_2020_CVPR,Tchapmi_2019_CVPR,lyu2021conditional,Xiang_2021_ICCV,zhou2022seedformer,zhu2023svdformer,xie2020grnet,zhang2020detail,Wang_2020_CVPR,wang2020cascaded,9525242,xiang2021snowflakenet,Tang_2022_CVPR,li2023proxyformer,Chen_2023_CVPR}. Though there are differences in feature extraction and utilization, the basic idea of these studies is to reconstruct the skeleton of the complete shape first and then refine it. PointTr \cite{yu2021pointr} does not follow the coarse-to-fine manner but only generates the missing part of the partial point cloud. The idea of only generating the missing part also appeared in \cite{ying2023adaptive,yu2023adapointr}. In \cite{sarmad2019rl}, a generative adversarial network is used for point cloud completion. PMP-Net \cite{Wen_2021_CVPR} and PMP-Net++ \cite{wen2022pmp} treat point cloud completion as a kind of deformation and complete point cloud by moving points to the right positions. P2C \cite{cui2023p2c} introduces additional losses to supervise the latent expressions. Other unsupervised point cloud completion works \cite{gong2022optimization,zhang2021unsupervised,hong2023acl,cao2023kt} achieve higher robustness through special training strategies Limited by inputs, these models must learn information about the complete shape from the occluded shape, which may lead to turning the task into a translation process from the occluded shape to the complete shape without meticulous analysis and design of the model.

\subsection{View-guided point cloud completion}
The purpose of the view-guided completion process is to introduce the missing geometric information from images to obtain better completion results. The pioneering work of View-guided point cloud completion is ViPC \cite{zhang2021view}, which designed a multi-modal architecture for image and point cloud and built the ShapeNet-ViPC dataset. The ViPC \cite{zhang2021view} model first used a modality transformer to convert images directly to skeleton point cloud and concatenate it with occluded point cloud, then refine it with concatenated image features and point cloud features. Rather than concatenation on results, CSDN \cite{zhu2023csdn} leverage the IPAdaIN \cite{huang2017arbitrary} to let image features affect the process of deforming point cloud features into coarse point clouds, then by using pixel-wise aligned local features to performer dual-refinement. XMFnet \cite{aiello2022cross} is the most recent baseline network that applied stacked cross-attention and self-attention layers to fuse the image feature with the point cloud feature and complete the point cloud in an end-to-end way using only the fused feature. The recent work CDPNet\cite{du2024cdpnet} introduce a two phase strategic which leverage global information from images to predict rough shape. These works show that the multi-modal information fusion strategy plays a critical role in view-guided point cloud completion.


%% file: Sec/method.tex
\section{Method}
\label{sec:method}
\begin{figure}
    \centering
    \includegraphics[width=1\linewidth]{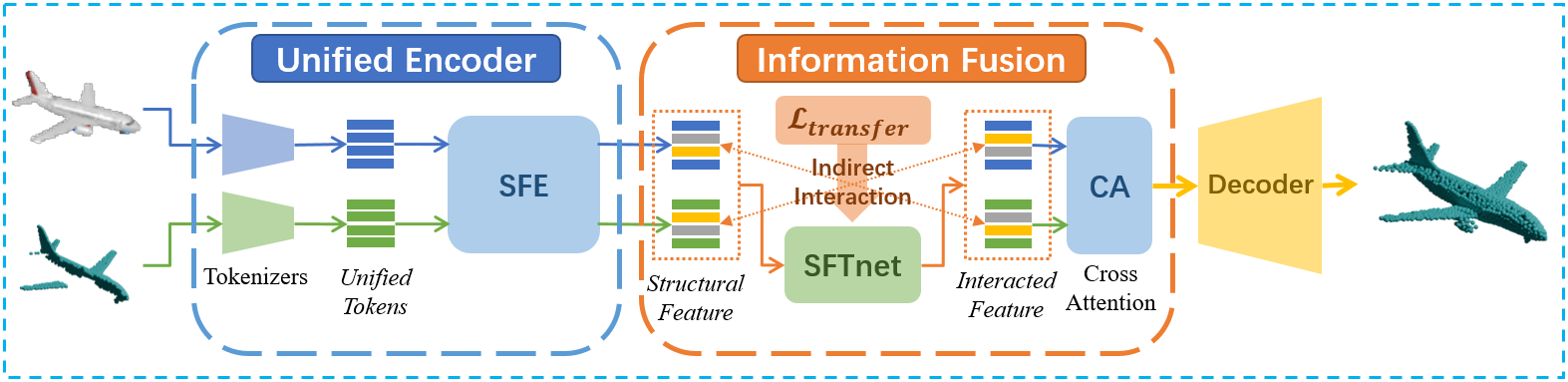}
    \caption{Architecture of EGIInet. The modal alignment is conducted by Unified Encoder. In the Information Fusion, FT-Loss ($\mathcal{L}_{transfer}$) explicitly guides the indirect interaction between image information and point cloud information.}
    \label{fig:2}
\end{figure}

The task of view-guided point cloud completion is to use an input occluded point cloud $\boldsymbol{P}\in\mathbb{R}^{N\times3}$  and a single view image $\boldsymbol{I}\in\mathbb{R}^{H\times W\times C}$ to predict the complete shape. The purpose of our design is to achieve better information fusion by performing modal alignment and information interaction, thus leading to a better prediction of the complete shape. To this purpose, we study i) a unified encoder for multi-modal input; and ii) a novel token feature structure transfer loss that guides the modal alignment and information interaction. Fig. \ref{fig:2} shows an overview of our proposed architecture EGIInet.

\subsection{Unified Encoder}
The main difficulty in designing feature fusion in multi-modal models is to overcome the domain gap between different modals \cite{pang2022masked}. Our design reduces modal differences in both format and latent space by utilizing tokenization techniques and shared structure. The proposed Unified Encoder consists of Tokenizers and a shared feature extractor (SFE) which will be detailed in the following.
\subsubsection{Tokenizers}
The gap between the image and point cloud lies in the differences in data organizing, so the first step of modal alignment is to give a unified way of describing the image and point cloud. Tokenization is a common technique to convert data into a sequence of tokens that is similar to the sentences in natural language. Therefore, tokens are ideal for uniform representation of images and point clouds since both of them can be described in natural language. By unifying the description of the image and point clouds, the following alignment in latent space can be simplified and information fusion can be conducted in a more explicit way. The function of tokenizers is to transfer point clouds $\boldsymbol{P}$ and images $\boldsymbol{I}$ into a unified format, that is features $\boldsymbol{F}\in\mathbb{R}^{N'\times C'}$ consists of $N'$ tokens $\boldsymbol{T}\in\mathbb{R}^{1\times C'}$. The point cloud feature $\boldsymbol{F}_{pc}$ and image feature $\boldsymbol{F}_{img}$ consist of same number($N'$) of tokens.

In order to explicitly guide the interaction of structural information, we need to first extract features that can represent both global structure and local geometry. For images, we can take advantage of the grid property of the image to represent the global structure using the organizational pattern among tokens. For point clouds, additional positional embedding is added to the token features to reduce the impact of the irregular nature of point clouds. Therefore, the image and point cloud tokenizers are designed to divide the image and the point cloud into several parts for mapping. In this way, the global structural information is contained in the organizational pattern among tokens and each token represents a certain local geometry.
\begin{figure}
    \centering
    \includegraphics[width=1\linewidth]{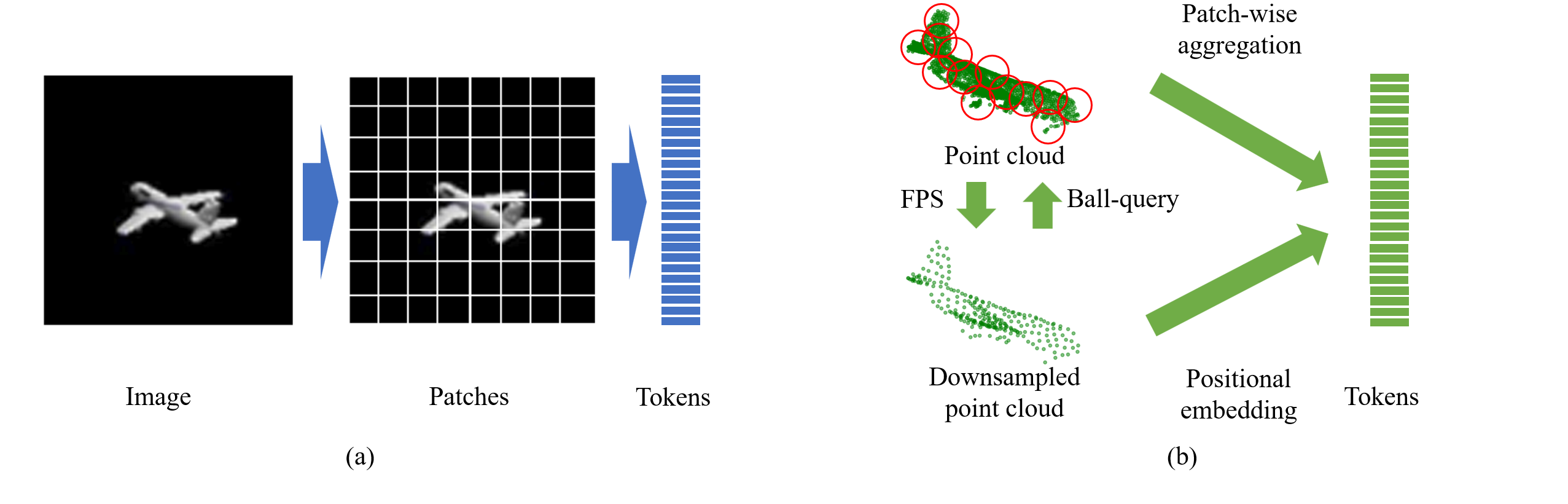}
    \caption{Tokenization process for images (a) and point clouds (b).}
    \label{fig:1.5}
\end{figure}
As shown in Fig. \ref{fig:1.5} (a), for tokenizing images, we use a convolution layer with large kernel size and stride to divide the image into several parts and each part is described by one token. In this way, we can learn a simple projection from image to tokens.

As shown in Fig. \ref{fig:1.5} (b), for tokenizing point clouds, we adopt $t$ steps of FPS (Farthest Point Sample) \cite{eldar1997farthest} downsampling while aggregating features in each step using Ball-query cluster. By aggregating the feature of each cluster, each point in the down-sampled point cloud $\boldsymbol{P}_{center}$ can be matched to one token and each token can describe the geometry of a specific area of the point cloud. In order to reduce the impact caused by the irregular nature of the point cloud, we extract the per-point features of the downsampled point cloud as the position embedding and add it to the tokens.

\subsubsection{Shared Feature Extractor (SFE)}
Another reason why multi-modal features are difficult to fuse is the difference between 2D backbones and 3D backbones. The features extracted by different network architectures have differences in latent distribution and semantic structure, which makes it difficult to merge information directly. To solve this problem, we use a unified shared architecture to learn the token sequences from different modals, so that the features of the image and point cloud are mapped to the adjacent latent space. Meanwhile, in order to make our model focus on the structural information that is essential for completion, we use self-attention-based ViT blocks \cite{dosovitskiy2020image} as the backbone of SFE. The SFE takes token sequence $\boldsymbol{F}_{pc},\boldsymbol{F}_{img}\in \mathbb{R}^{N'\times C'}$ as input, and export processed features $\boldsymbol{F}_{pc}^{stc},\boldsymbol{F}_{img}^{stc}\in \mathbb{R}^{N'\times C'}$. The process of SFE can be described in Formula \ref{form.1} and Formula \ref{form.2}.
\begin{eqnarray}
\label{form.1}
   \boldsymbol{F}_{pc}^{stc}=\text{SFE}(\boldsymbol{F}_{pc}) \\
\label{form.2}
   \boldsymbol{F}_{img}^{stc}=\text{SFE}(\boldsymbol{F}_{img}) 
\end{eqnarray}
\subsection{Information Fusion}
Intuitively, the most critical information contained in the image from a different sight is that representing the missing part of the point cloud. However, the traditional latent fusion strategy could not always focus on this critical information due to the lack of structural guidance, leading to a sub-optimal solution for point cloud completion. To effectively fuse the critical information into the inference process of missing part, we introduce a dual-designed loss function to explicitly guide an information interaction process separated from the encoding stage. In the following sections we introduce the Shared Feature Transfer Network (SFTnet) which provides the information interaction process and the Feature Transfer Loss (FT-Loss) which explicitly guides the information interaction.
\subsubsection{Shared Feature Transfer Network (SFTnet)}
\label{sft}
Separating the information interaction process from the encoding process makes the network have specific learning objectives at specific stages, thereby reducing the overall optimization difficulty. Meanwhile, we observe that directly fusing the point cloud and image features in latent space like previous methods will lead to an ambiguous feature interaction, as there is no explicit guidance to decide which part of the image contains critical information. Also, direct feature fusion will change the organizational pattern of features, leading to an extra learning on new latent expressions. Therefore, the SFTnet is purposed to give an independent interaction process without direct contact between features. In this way, point cloud features and image features can interact with each other in an explicitly guided manner while maintaining their respective information organization pattern. This transfer process is supervised by the Feature Transfer Loss (FT-Loss) which will be detailed in the next section.

The SFTnet consists of ViT-based blocks \cite{dosovitskiy2020image} similar to SFE in implementation. The reason for using the similar design of SFE is that a unified shared design helps to maintain the information organization pattern of the features, so as to conduct the information interaction without destroying the original structure of the features. The process of Shared Feature Transfer Network can be described in Formula \ref{form.3} and Formula \ref{form.4}.
\begin{eqnarray}
\label{form.3}
   \boldsymbol{F}_{pc}'=\text{SFTnet}(\boldsymbol{F}_{pc}^{stc}) \\
\label{form.4}
   \boldsymbol{F}_{img}'=\text{SFTnet}(\boldsymbol{F}_{img}^{stc}) 
\end{eqnarray}
\subsubsection{Feature Transfer Loss (FT-Loss)}
\label{ft-l}
We implement explicit guidance for feature interaction in the form of loss supervision, in this way we can artificially determine the information that features need to be interacted with. We explicitly guide the information transfer between image features and point cloud features to conduct the identifying of the critical information within the image and the transformation of the critical information from the image features to the point cloud features, thus achieving the ultimate goal of making the critical information in the image act in the point cloud completion. 


The purposed FT-Loss $\mathcal{L}_{transfer}$ consists of Informational Loss $\mathcal{L}_{infor}$ and Structural Loss $\mathcal{L}_{stc}$. The function of $\mathcal{L}_{infor}$ is to interact with the critical structural information in image features and point cloud features while the function of $\mathcal{L}_{stc}$ is to maintain the information structure of the point cloud features.

We leverage the Gram matrix of features as the basis for information loss since the Gram matrix provides a way to describe the structural criticality of features. The Gram matrix can be considered as an eccentricity covariance matrix for features and can be calculated through Formula \ref{form.5}. For each feature, each element of its Gram matrix corresponds to a channel-wise global structural criticality.
\begin{eqnarray}
\label{form.5}
\boldsymbol{G}\left(\boldsymbol{F}\right)=\boldsymbol{F}^T\bullet \boldsymbol{F}
\end{eqnarray}

The purpose of Information Loss $\mathcal{L}_{infor}$ is to make the features of one modal perceive the structural information present in the features of another modal. To achieve this, we adopt a dual-designed loss to pass information between the transferring processes of images and point clouds. The Information Loss $\mathcal{L}_{infor}$ is defined as Formula \ref{form.6}. By supervising the similarity of the Gram matrix of features, we can indirectly align the structural criticality of features, thus achieving structural information transformation.  Through the alignment of structural criticality, the missing relationship contained in point cloud features can be transferred to the image features and the structure of the missing part contained in image features can be transferred to the point cloud features.
\begin{eqnarray}
\label{form.6}
\mathcal{L}_{infor}=\frac{\left(\boldsymbol{G}\left(\boldsymbol{F}_{img}^{stc}\right)-\boldsymbol{G}\left(\boldsymbol{F}_{pc}'\right)\right)^2+\left(\boldsymbol{G}\left(\boldsymbol{F}_{pc}^{stc}\right)-\boldsymbol{G}\left(\boldsymbol{F}_{img}'\right)\right)^2}{N\times C}
\end{eqnarray}

Where $\boldsymbol{F}_{img}^{stc}$, $\boldsymbol{F}_{pc}^{stc}$ are inputs of SFTnet and $\boldsymbol{F}_{img}'$, $\boldsymbol{F}_{pc}'$ are outputs of SFTnet.

As mentioned before, 2D features from images have difficulty predicting 3D coordinates directly, so the fused feature used to reconstruct the missing part should be based on 3D features from point clouds. The purpose of Structural Loss is to maintain the information structure of 3D point cloud features through the transfer process. The Structural Loss $\mathcal{L}_{stc}$ is defined as Formula \ref{form.7}.
\begin{eqnarray}
\label{form.7}
    \mathcal{L}_{stc}=\left(\boldsymbol{F}_{pc}^{stc}-\boldsymbol{F}_{pc}'\right)^2
\end{eqnarray}

By supervising these two losses simultaneously we provide an explicitly guided information interaction method to transfer information between to modals, thus improving the fusion efficiency. The FT-Loss $\mathcal{L}_{transfer}$ is defined as Formula \ref{form.8}.
\begin{eqnarray}
\label{form.8}
    \mathcal{L}_{transfer}=\mathcal{L}_{infor}+\mathcal{L}_{stc}
\end{eqnarray}
Chamfer Distance (CD)\cite{yang2019pointflow} is widely used in the reconstruction task. The calculation of $l_1-\text{CD}$ is shown in Formula \ref{form.9} where $\boldsymbol{P}=\left\{p\in\mathbb{R}^3\right\}$ is the ground truth point cloud and $\boldsymbol{\hat{P}}=\left\{\hat{p}\in\mathbb{R}^3\right\}$ is the output completed point cloud of our model.
\begin{eqnarray}
\label{form.9}
    \mathcal{L}_{l_1-\text{CD}}\left(\boldsymbol{P},\boldsymbol{\hat{P}}\right)=\frac{1}{2N}\sum_{p\in \boldsymbol{P}}\min_{\hat{p}\in\boldsymbol{\hat{P}}}||p-\hat{p}||_2+\frac{1}{2N}\sum_{\hat{p}\in \boldsymbol{\hat{P}}}\min_{p\in \boldsymbol{P}}||p-\hat{p}||_2
\end{eqnarray}

Together with Chamfer Distance, the total loss of our architecture can be defined as Formula \ref{form.10} where $\alpha$ is a hyperparameter. In implement, $\alpha$ is fix to 0.01 since $\mathcal{L}_{transfer}$ is a large value compared with $\mathcal{L}_{l_1-CD}$
\begin{eqnarray}
\label{form.10}
\mathcal{L}_{total}=\alpha\times\mathcal{L}_{transfer}+\mathcal{L}_{l_1-CD}
\end{eqnarray}
\subsubsection{Feature Fusion}
To aggregate the features, we adopt a simple cross-attention layer to fuse the image feature and point cloud feature since these two features have been fully interacted within the previous process.
\subsection{Completion Decoder}
In order to decode the acquired fusion features into the complete point cloud, we need a decoder that is flexible and has a certain learning ability. To do this, we use a decoder architecture similar to XMFnet \cite{aiello2022cross} to accept similar fused features and learn their implicit expressions to predict 3D coordinates.

%% file: Sec/experiment.tex
\section{Experimental Results}
\label{sec:experiment}
In this section, we first introduce the dataset and evaluation metrics in section \ref{sec:4.1}. Quantitative and Qualitative comparisons are shown in section \ref{sec:4.2}. Ablation studies are conducted in section \ref{sec:4.3}. We also report the generalization ability of our method in section \ref{sec:4.4}. Finally, the model complexity can be found in supplementary materials.
\subsection{Experimental Settings}
\label{sec:4.1}
\subsubsection{Dataset} In this work we train and test our model on ShapeNet-ViPC dataset \cite{zhang2021view}. The dataset contains 38,328 objects from 13 categories. For each object, ViPC \cite{zhang2021view} generates 24 incomplete point clouds under 24 viewpoints. In this paper, we follow the same dataset settings of ViPC \cite{zhang2021view}.

\subsubsection{Evaluation metrics} We use $l_2-\text{CD}$ \cite{yang2019pointflow} and F-score \cite{knapitsch2017tanks} to evaluate our model the same as the previous works do. The $l_2-\text{CD}$ of point cloud $\boldsymbol{X}$ and $\boldsymbol{Y}$ is calculated as shown in Formula \ref{form.11}, where $N_X$ and $N_Y$ denotes the number of points in $\boldsymbol{X}$ and $\boldsymbol{Y}$. Since CDPNet \cite{du2024cdpnet} was not open sourced during our study, we did not compare with it.
\begin{eqnarray}
\label{form.11}
        \text{CD}\left(\boldsymbol{X},\boldsymbol{Y}\right)=\frac{1}{N_X}\sum_{x\in \boldsymbol{X}}\min_{y\in \boldsymbol{Y}}||x-y||_2^2+\frac{1}{N_Y}\sum_{y\in \boldsymbol{Y}}\min_{x\in \boldsymbol{X}}||x-y||_2^2
\end{eqnarray}
The F-score \cite{knapitsch2017tanks} is defined in Formula \ref{form.12}, where thresh hold $d$ equals 0.001 the same as the previous works.
\begin{eqnarray}
\label{form.12}
        \text{F}(X,Y)=\frac{2X(d)Y(d)}{X(d)+Y(d)},
\end{eqnarray}
where $X(d)$ and $Y(d)$ denote the mean of the squared distances that are less than the threshold $d$. The calculation of squared distances follows the same in calculating CD.
\subsection{Results on ShapeNet-ViPC}
\label{sec:4.2}
\begin{figure}
    \centering
    \includegraphics[width=1\linewidth]{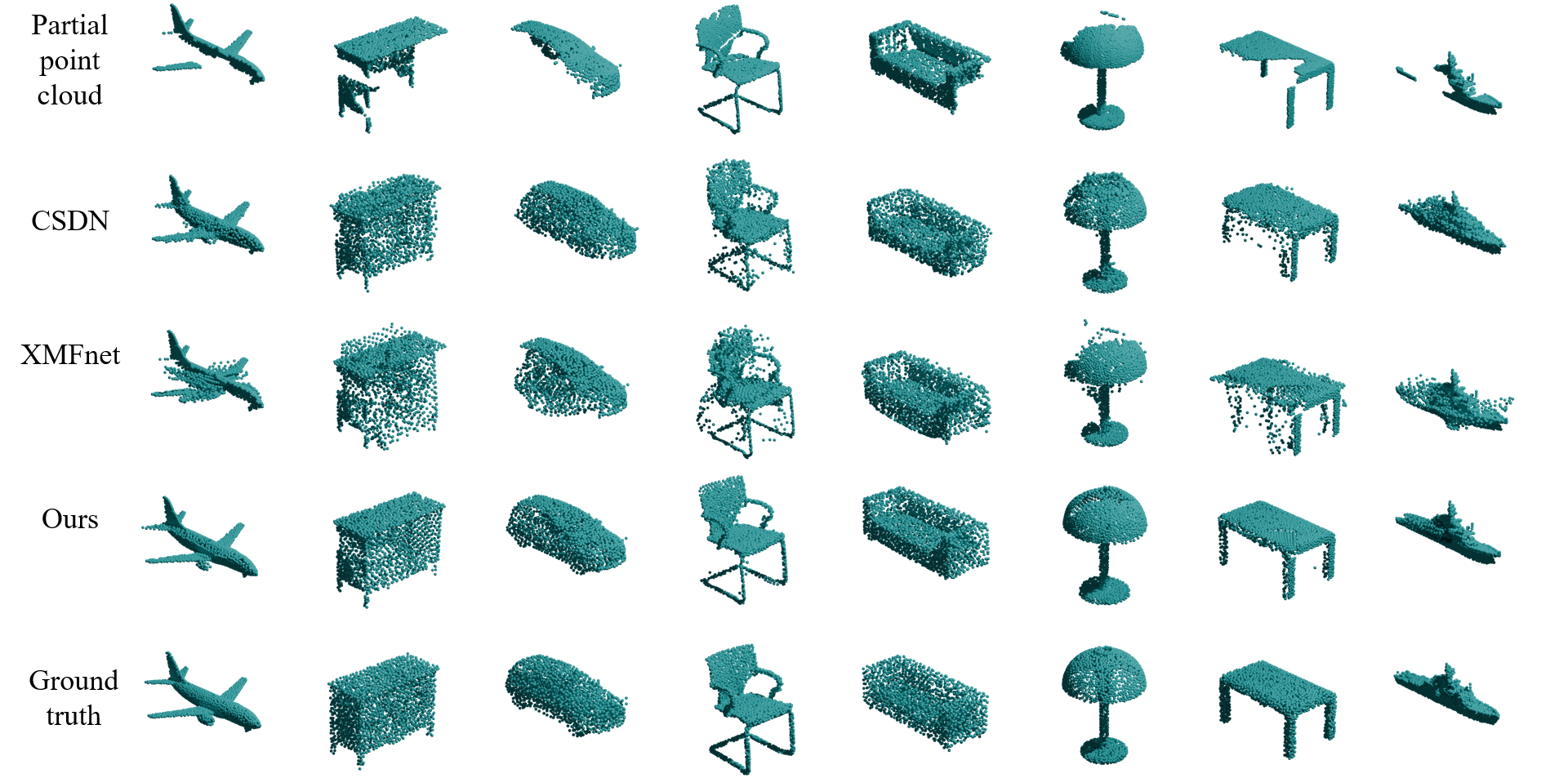}
    \caption{Qualitative results on ShapeNet-ViPC dataset \cite{zhang2021view}.}
    \label{fig:3}
\end{figure}

In this section we report our quantitative comparison with the existing works \cite{zhang2021view,zhu2023csdn,aiello2022cross} that use the same training data on ShapeNet-ViPC dataset \cite{zhang2021view} and other SOTA point cloud completion methods \cite{groueix2018papier,yang2017foldingnet,yuan2018pcn,Tchapmi_2019_CVPR,Huang_2020_CVPR,liu2020morphing,xie2020grnet,yu2021pointr,wang2022pointattn,zhang2022point,zhou2022seedformer} taken from \cite{zhu2023csdn}, results are reported in Table \ref{table:1} for the CD and Table \ref{table:2} for the F1-score. We conduct a qualitative comparison with CSDN \cite{zhu2023csdn} and XMFnet \cite{aiello2022cross} in Fig. \ref{fig:3}. It is shown that while keeping the decoding structure the same, our model is able to improve the CD by 16\% compared to XMFnet \cite{aiello2022cross}. This means that our design of better feature extraction and fusion can lead to better completion results. Specifically we achieve a great improve on lamps due to the accurately extracted information and well designed information interaction.
\begin{table}[h]
  \begin{center}
  \caption{Mean Chamfer Distance per point ($\text{CD}\times 10^3\downarrow$).}
    \begin{tabular}{l|c c c c c c c c c}
    \hline
    Methods&Avg&Airplane&Cabinet&Car&Chair&Lamp&Sofa&Table&Watercraft \\
    \hline
    AtlasNet\cite{groueix2018papier}&6.062&5.032&6.414&4.868&8.161&7.182&6.023&6.561&4.261 \\
    FoldingNet\cite{yang2017foldingnet}&6.271&5.242&6.958&5.307&8.823&6.504&6.368&7.080&3.882 \\
    PCN\cite{yuan2018pcn}&5.619&4.246&6.409&4.840&7.441&6.331&5.668&6.508&3.510 \\
    TopNet\cite{Tchapmi_2019_CVPR}&4.976&3.710&5.629&4.530&6.391&5.547&5.281&5.381&3.350 \\
    PF-Net\cite{Huang_2020_CVPR}&3.873&2.515&4.453&3.602&4.478&5.185&4.113&3.838&2.871 \\
    MSN\cite{liu2020morphing}&3.793&2.038&5.060&4.322&4.135&4.247&4.183&3.976&2.379 \\
    GRNet\cite{xie2020grnet}&3.171&1.916&4.468&3.915&3.402&3.034&3.872&3.071&2.160 \\
    PoinTr\cite{yu2021pointr}&2.851&1.686&4.001&3.203&3.111&2.928&3.507&2.845&1.737 \\
    PointAttN\cite{wang2022pointattn}&2.853&1.613&3.969&3.257&3.157&3.058&3.406&2.787&1.872 \\
    SDT\cite{zhang2022point}&4.246&3.166&4.807&3.607&5.056&6.101&4.525&3.995&2.856 \\
    Seedformer\cite{zhou2022seedformer}&2.902&1.716&4.049&3.392&3.151&3.226&3.603&2.803&1.679 \\
    \hline
    ViPC\cite{zhang2021view}&3.308&1.760&4.558&3.183&2.476&2.867&4.481&4.990&2.197 \\
    CSDN\cite{zhu2023csdn}&2.570&1.251&3.670&2.977&2.835&2.554&3.240&2.575&1.742 \\
    XMFnet\cite{aiello2022cross}&1.443&0.572&1.980&1.754&1.403&1.810&1.702&1.386&0.945 \\
    \hline
    Ours&\textbf{1.211}&\textbf{0.534}&\textbf{1.921}&\textbf{1.655}&\textbf{1.204}&\textbf{0.776}&\textbf{1.552}&\textbf{1.227}&\textbf{0.802} \\
    \hline
    \end{tabular}
    
  \label{table:1}%
  \end{center}
\end{table}

\begin{table}[h]
  \begin{center}
  \caption{Mean F-Score @ 0.001$\uparrow$.}
    \begin{tabular}{l|c c c c c c c c c}
    \hline
    Methods&Avg&Airplane&Cabinet&Car&Chair&Lamp&Sofa&Table&Watercraft \\
    \hline
    AtlasNet\cite{groueix2018papier}&0.410&0.509&0.304&0.379&0.326&0.426&0.318&0.469&0.551 \\
    FoldingNet\cite{yang2017foldingnet}&0.331&0.432&0.237&0.300&0.204&0.360&0.249&0.351&0.518 \\
    PCN\cite{yuan2018pcn}&0.407&0.578&0.270&0.331&0.323&0.456&0.293&0.431&0.577 \\
    TopNet\cite{Tchapmi_2019_CVPR}&0.467&0.593&0.358&0.405&0.388&0.491&0.361&0.528&0.615 \\
    PF-Net\cite{Huang_2020_CVPR}&0.551&0.718&0.399&0.453&0.489&0.559&0.409&0.614&0.656 \\
    MSN\cite{liu2020morphing}&0.578&0.798&0.378&0.380&0.562&0.652&0.410&0.615&0.708 \\
    GRNet\cite{xie2020grnet}&0.601&0.767&0.426&0.446&0.575&0.694&0.450&0.639&0.704 \\
    PoinTr\cite{yu2021pointr}&0.683&0.842&0.516&0.545&0.662&0.742&0.547&0.723&0.780 \\
    PointAttN\cite{wang2022pointattn}&0.662&0.841&0.483&0.515&0.638&0.729&0.512&0.699&0.774 \\
    SDT\cite{zhang2022point}&0.473&0.636&0.291&0.363&0.398&0.442&0.307&0.574&0.602 \\
    Seedformer\cite{zhou2022seedformer}&0.688&0.835&0.551&0.544&0.668&0.777&0.555&0.716&0.786 \\
    \hline
    ViPC\cite{zhang2021view}&0.591&0.803&0.451&0.512&0.529&0.706&0.434&0.594&0.730 \\
    CSDN\cite{zhu2023csdn}&0.695&0.862&0.548&0.560&0.669&0.761&0.557&0.729&0.782 \\
    XMFnet\cite{aiello2022cross}&0.796&0.961&0.662&0.691&0.809&0.792&0.723&0.830&0.901 \\
    \hline
    Ours&\textbf{0.836}&\textbf{0.969}&\textbf{0.693}&\textbf{0.723}&\textbf{0.847}&\textbf{0.919}&\textbf{0.756}&\textbf{0.857}&\textbf{0.927} \\
    \hline
    \end{tabular}
    
  \label{table:2}%
  \end{center}
\end{table}
\subsection{Ablation Studies}
\label{sec:4.3}
We first report the ablation on our model components. The object of ablation is the shared structure, FT-Loss and SFTnet. Then we analysis about the efficiency of image information.
\subsubsection{Ablation on FT-Loss}
\begin{figure}
    \centering
    \includegraphics[width=0.8\linewidth]{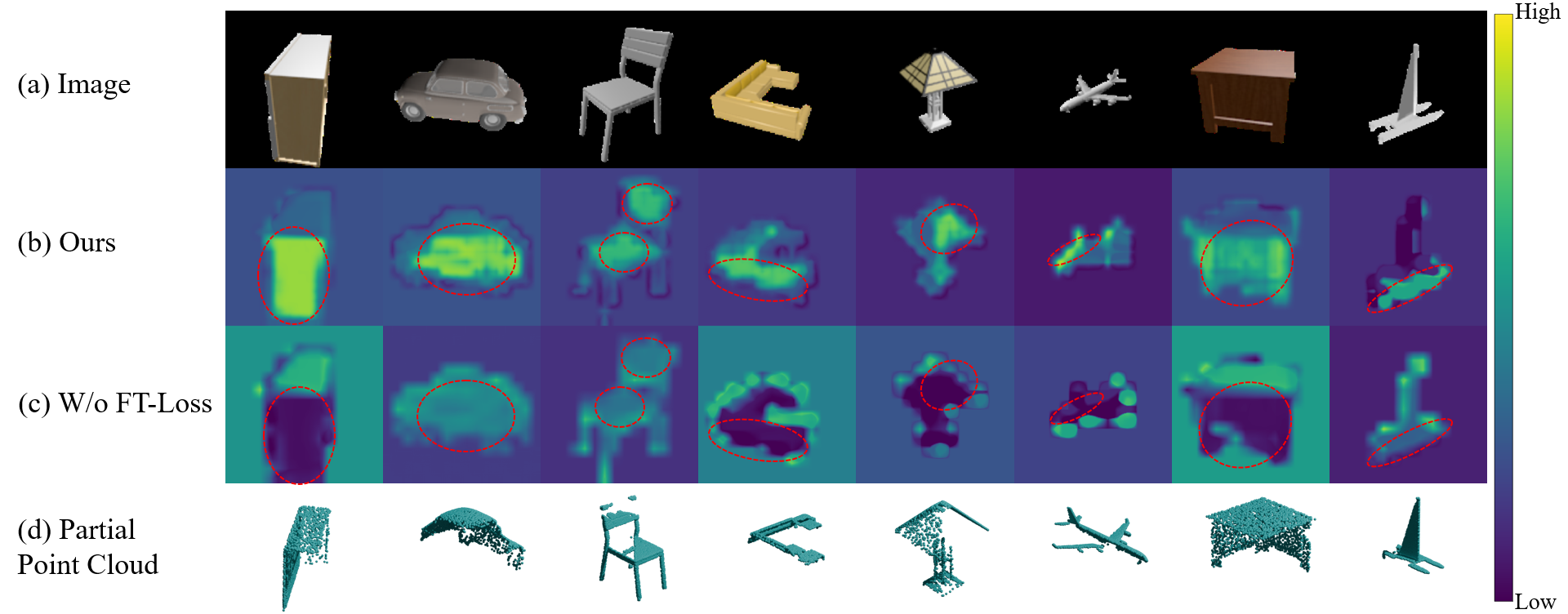}
    \caption{Image feature projection of our method with (b) and without (c) FT-loss. By supervising the FT-Loss, information interaction can be explicitly guided so that the image feature can figure out the most helpful information for point cloud completion.}
    \label{fig:4}
\end{figure}

To verify the effectiveness of FT-Loss, we do not calculate and supervise the loss during the process of training. Results are presented in Table \ref{table:3}. It can be seen that removing FT-Loss will decline the model's performance, and the missing relationship can not be transferred to the image feature without supervising the FT-Loss,  as shown in Fig. \ref{fig:4}. 
\subsubsection{Ablation on shared structure}
To verify the effectiveness of the shared structure, we replicate the shared ViT \cite{dosovitskiy2020image} blocks, including SFE and SFTnet, and pass the image and point cloud tokens through separate networks. Results are presented in Table \ref{table:3}. The CD metric decreased significantly without the shared structure, indicating that the modal alignment achieved by the shared structure can promote subsequent interaction and fusion. Due to the limited parameters, the full model performs slightly worse than the model without shared encoders on some complex classes (less valid pixels). However, for most of the common categories, the shared structure can effectively align different modalities with fewer parameters.

\begin{table}[h]
  \begin{center}
  \setlength{\abovecaptionskip}{0.1cm}
  \setlength{\belowcaptionskip}{0.1cm}
  \caption{Results of ablation studies ($\text{CD}\times 10^3\downarrow$).}
    \begin{tabular}{l|c c c c c c c c c}
    \hline
    Methods&Avg&Airplane&Cabinet&Car&Chair&Lamp&Sofa&Table&Watercraft \\
    \hline    Ours&\textbf{1.211}&\textbf{0.534}&\textbf{1.921}&\textbf{1.655}&\textbf{1.204}&0.776&\textbf{1.552}&\textbf{1.227}&0.802 \\
    w/o sharing&1.429&0.631&2.027&2.112&1.701&\textbf{0.711}&2.031&1.440&\textbf{0.783} \\
    w/o FT-Loss&1.354&0.580&2.137&1.911&1.288&0.824&1.732&1.484&0.875 \\
    w/o SFTnet&1.454&0.656&2.329&2.106&1.460&0.839&1.938&1.395&0.907 \\
    w/o image&1.383&0.583&2.110&1.953&1.343&0.860&1.830&1.510&0.874 \\
    
    \hline 
    \end{tabular}
    
  \label{table:3}%
  \end{center}
\end{table}
\subsubsection{Ablation on SFTnet}
In order to verify the necessity of separating the information interaction process from the encoding process, we designed an ablation experiment on STFnet. In this experiment, we remove SFTnet and calculate only the direct information loss of structural features shown as Formula \ref{form.13}. As shown in Table \ref{table:3}, it is difficult to complete the feature extraction and information interaction by relying only on SFE and simplified losses $\mathcal{L}_{transfer}'$, especially on some complex categories (lamp, watercraft, etc.). The reason behind this is that SFTnet can provide a more effective interaction for completion.
\begin{eqnarray}
\label{form.13}
\mathcal{L}_{transfer}'=\frac{\left(\boldsymbol{G}\left(\boldsymbol{F}_{img}^{stc}\right)-\boldsymbol{G}\left(\boldsymbol{F}_{pc}^{str}\right)\right)^2}{N\times C}
\end{eqnarray}
\subsubsection{Ablation of Input Modality}
To verify the effectiveness of input images, we only use point cloud as input, thus verifying that our design is able to make the information provided by the image positive. Results are presented in Table \ref{table:3}. 

\subsubsection{Efficiency of Image Information}
\begin{figure}
    \centering
    \includegraphics[width=0.8\linewidth]{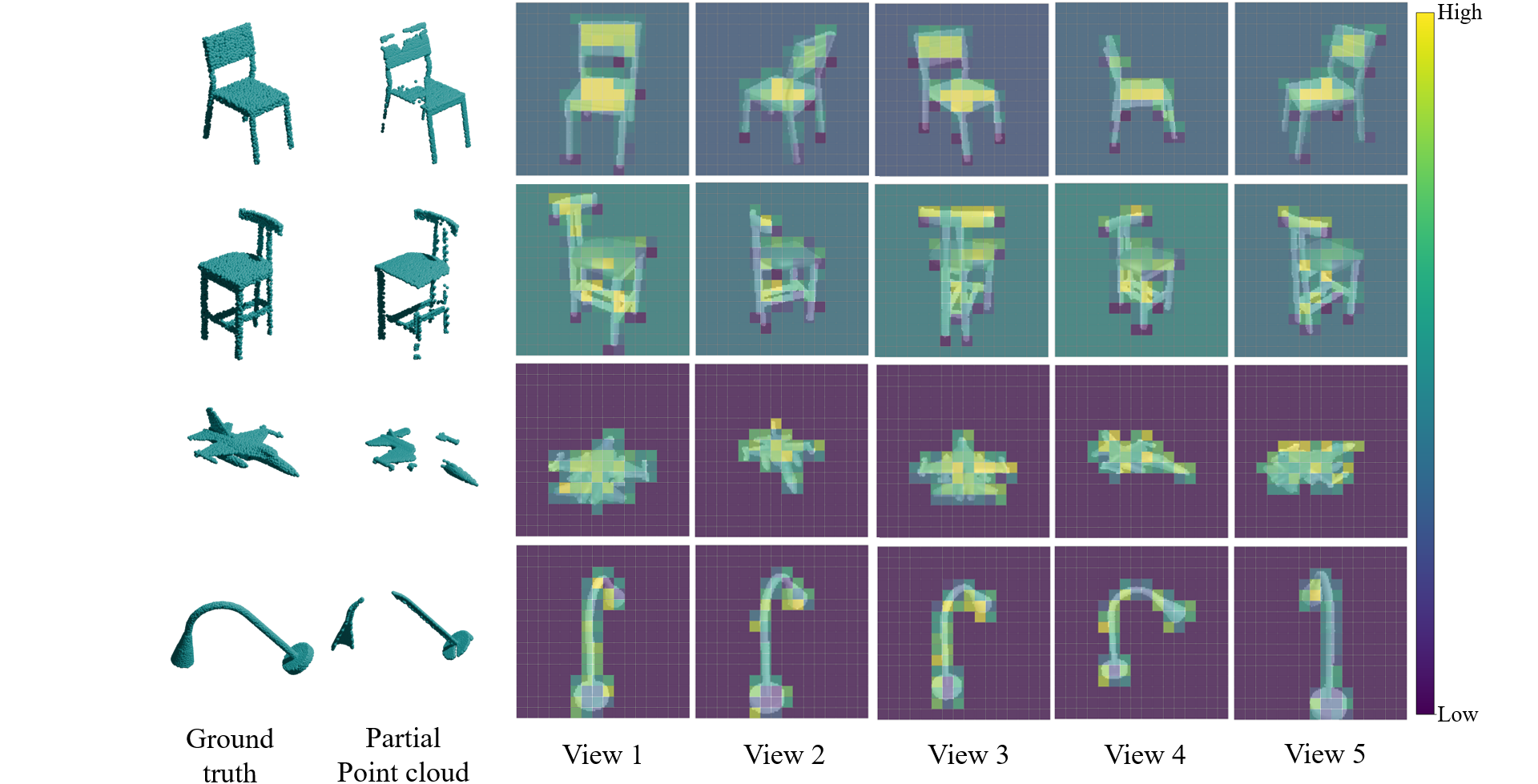}
    \caption{Cross-attention weight map projection of different views.}
    \label{fig:7}
\end{figure}

The projection of the cross-attention weight map plot in Fig. \ref{fig:7} shows that the image features of different views are able to focus on the geometer related to the missing part of the point cloud, thus proving the effectiveness of information interaction. 

\subsection{Generalization Ability Evaluation}
\label{sec:4.4}
\subsubsection{Results on Unknown categories of ShapeNet-ViPC}
\begin{figure}
    \centering
    \includegraphics[width=0.9\linewidth]{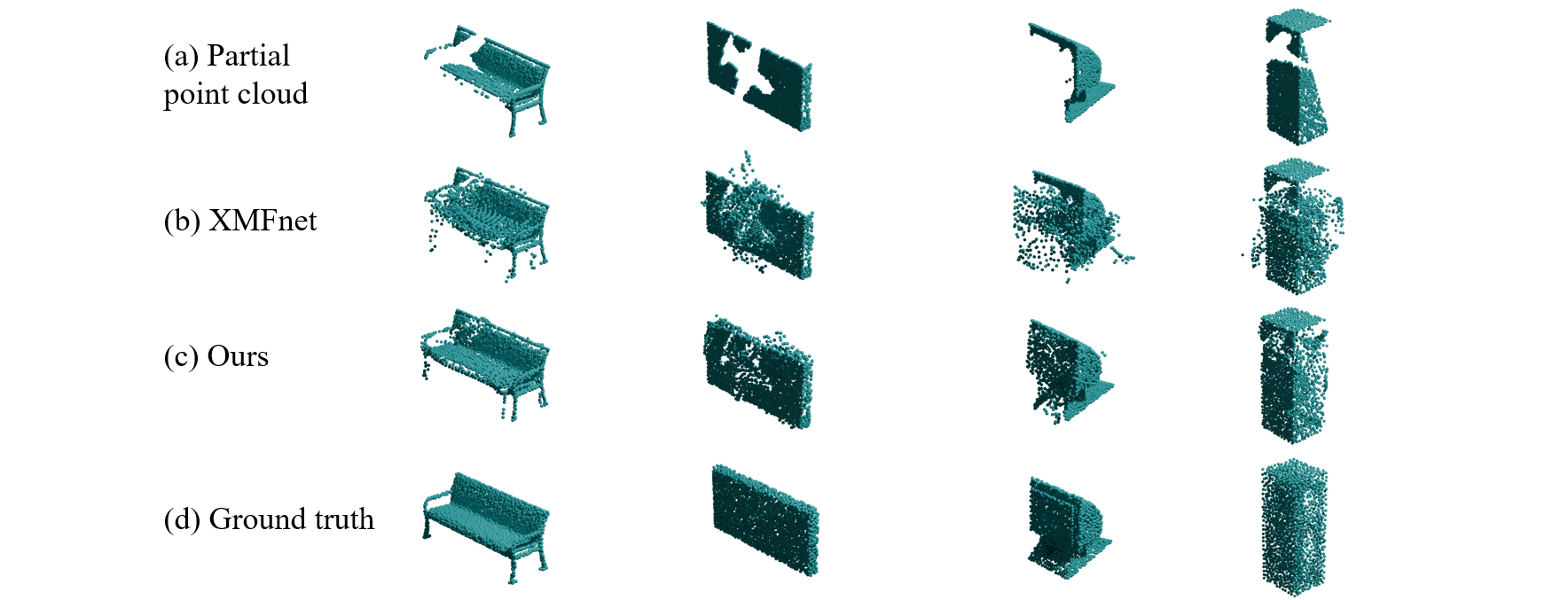}
    \caption{Qualitative results on unknown categories of ShapeNet-ViPC\cite{zhang2021view}.}
    \label{fig:5}
\end{figure}

To verify the utility of our method, we conducted a zero-shot test on unknown categories in the ShapeNet-ViPC dataset. We used 8 known categories, including airplane, cabinet, car, chair, lamp, sofa, table, and watercraft in the training stage, and tested on 4 unknown categories, including bench, monitor, speaker, and cellphone. We compare the CD and F-score performance on 4 categories with other method \cite{Huang_2020_CVPR,liu2020morphing,xie2020grnet,yu2021pointr,wang2022pointattn,zhang2022point} taken from \cite{zhu2023csdn} and we train XMFnet\cite{aiello2022cross} on 8 categories as well. Quantitative comparisons are shown in table \ref{table:5} and qualitative comparisons are shown in Fig. \ref{fig:5}.

\begin{table}
  \begin{center}
  \caption{Results on unknown categories ($\text{CD}\times 10^3\downarrow$, F-score @ 0.001$\uparrow$).}
    \begin{tabular}{l|c c|c c|c c|c c|c c}
    \hline
    
    \multirow{2}{*}{Methods}&\multicolumn{2}{c|}{Avg}&\multicolumn{2}{c|}{Bench}&\multicolumn{2}{c|}{Monitor}&\multicolumn{2}{c|}{Speaker}&\multicolumn{2}{c}{cellphone} \\
    ~&CD&F-score&CD&F-score&CD&F-score&CD&F-score&CD&F-score \\
    \hline
    PF-Net\cite{Huang_2020_CVPR}&5.011&0.468&3.683&0.584&5.304&0.433&7.663&0.319&3.392&0.534 \\ MSN\cite{liu2020morphing}&4.684&0.533&2.613&0.706&4.818&0.527&8.259&0.291&3.047&0.607 \\
    GRNet\cite{xie2020grnet}&4.096&0.548&2.367&0.711&4.102&0.537&6.493&0.376&3.422&0.569 \\
    PoinTr\cite{yu2021pointr}&3.755&0.619&1.976&0.797&4.084&0.599&5.913&0.454&3.049&0.627 \\
    PointAttN\cite{wang2022pointattn}&3.674&0.605&2.135&0.764&3.741&0.591&5.973&0.428&2.848&0.637 \\
    SDT\cite{zhang2022point}&6.001&0.327&4.096&0.479&6.222&0.268&9.499&0.197&4.189&0.362 \\
    \hline
    ViPC\cite{zhang2021view}&4.601&0.498&3.091&0.654&4.419&0.491&7.674&0.313&3.219&0.535 \\
    CSDN\cite{zhu2023csdn}&3.656&0.631&1.834&0.798&4.115&0.598&5.690&0.485&2.985&0.644 \\
    XMFnet\cite{aiello2022cross}&2.671&0.710&1.278&0.862&2.806&0.677&4.823&0.556&1.779&0.748 \\
    \hline
    Ours&\textbf{2.354}&\textbf{0.750}&\textbf{1.047}&\textbf{0.902}&\textbf{2.513}&\textbf{0.716}&\textbf{4.282}&\textbf{0.591}&\textbf{1.575}&\textbf{0.792} \\
    \hline 
    \end{tabular}
    
  \label{table:5}%
  \end{center}
\end{table}


\subsubsection{Results on Real Scenes}
We also report qualitative results on KITTI \cite{geiger2013vision} cars extracted by \cite{yuan2018pcn}. Results about qualitative results can be found in supplementary materials.


    \label{fig:6}

%% file: Sec/conclusion.tex
\section{Conclusions}
\label{sec:conclusion}
In this paper, we propose an explicitly guided information interaction strategy supported by modal alignment for view-guided point cloud completion. This explicit guidance can promote the network to learn structural relationships for completion, thus leading to better utilization of the information provided by the image. Our proposed methods achieve new SOTA results on the ShapeNet-ViPC dataset \cite{zhang2021view}. 
In future work, we will continue to study this information fusion approach and have the potential to extend it to other data modalities and tasks to make it a new multi-modal learning paradigm.